\documentclass[a4paper]{article}

\usepackage{INTERSPEECH2018}
\usepackage{multirow}
\usepackage{hhline}
\usepackage{subfigure}
\usepackage{url}
\usepackage{comment}
\usepackage{color}

\title{Speech2Vec: A Sequence-to-Sequence Framework\\for Learning Word Embeddings from Speech}
\name{Yu-An Chung\quad James Glass}

\address{
  Computer Science and Artificial Intelligence Laboratory\\
  Massachusetts Institute of Technology\\
  Cambridge, MA 02139, USA}
\email{\{andyyuan,glass\}@mit.edu}

\begin{document}

\maketitle
 
\begin{abstract}

In this paper, we propose a novel deep neural network architecture, Speech2Vec, for learning fixed-length vector representations of audio segments excised from a speech corpus, where the vectors contain semantic information pertaining to the underlying spoken words, and are close to other vectors in the embedding space if their corresponding underlying spoken words are semantically similar.
The proposed model can be viewed as a speech version of Word2Vec~\cite{mikolov2013distributed}.
Its design is based on a RNN Encoder-Decoder framework, and borrows the methodology of skipgrams or continuous bag-of-words for training.
Learning word embeddings directly from speech enables Speech2Vec to make use of the semantic information carried by speech that does not exist in plain text.
The learned word embeddings are evaluated and analyzed on 13 widely used word similarity benchmarks, and outperform word embeddings learned by Word2Vec from the transcriptions.

\end{abstract}
\noindent\textbf{Index Terms}: word embeddings, recurrent neural networks, sequence-to-sequence learning, skipgrams, bag-of-words

\section{Introduction}
Natural language processing~(NLP) techniques such as Word2Vec~\cite{mikolov2013distributed,bojanowski2017enriching} and GloVe~\cite{pennington2014glove} transform words into fixed dimensional vectors, or word embeddings.
The embeddings are obtained via unsupervised learning from co-occurrence information in text, and contain semantic information about the word which are useful for many NLP tasks~\cite{yu2017character,lample2016neural,plank2016multilingual,kim2016character,ballesteros2015improved}.

Researchers have also explored the concept of learning vector representations from speech~\cite{he2017multi,settle2016discriminative,chung2016audio,kamper2016deep,bengio2014word,levin2013fixed}.
These approaches are based on notions of acoustic-phonetic~(rather than {\it semantic}) similarity, so that different instances of the same underlying word would map to the same point in a latent embedding space.
Our work, highly inspired by Word2Vec~\cite{mikolov2013distributed}, uses a skipgrams or continuous bag-of-words formulation to focus on {\it neighboring} acoustic regions, rather than the acoustic segment associated with the word itself.
We show that the resulting acoustic embedding space is more semantic in nature.

Recent research by~\cite{harwath2017learning,harwath2016unsupervised,harwath2015deep} has presented a deep neural network model capable of rudimentary spoken language acquisition using raw speech training data paired with contextually relevant images.
Using this contextual grounding, the model learned a latent semantic audio-visual embedding space.
In this paper, we propose a deep neural network architecture capable of learning embeddings of audio segments corresponding to words from \textit{raw} speech without any other modalities.
The proposed model, called Speech2Vec, integrates an RNN Encoder-Decoder framework~\cite{sutskever2014sequence,cho2014learning} with the concept of skipgrams or continuous bag-of-words, and can handle arbitrary length audio segments.
The resulting word embeddings contain information pertaining to the meaning of the underlying spoken words such that semantically similar words produce vector representations that are nearby in the embedding space.

Speech2Vec can be viewed as a speech version of Word2Vec.
Traditionally, when we want to learn word embeddings from speech, we need to first transcribe the speech into text by an ASR system, then apply a textual word embedding method on the transcripts.
The motivations for this work are that learning word embeddings directly from speech surmounts the recognition errors caused by the process of transcribing.
Moreover, speech contains richer information than text such as prosody, and a machine should be able to make use of this information in order to learn better semantic representations.

In this paper, we build on a preliminary version of the Speech2Vec model~\cite{chung2017learning} by introducing additional methodologies and details for training the model, comparing the model with additional baseline approaches, and providing systematic analysis and visualizations of the learned word embeddings.

\section{Proposed Approach}
Our goal is to learn a fixed-length embedding of a audio segment corresponding to a word that is represented by a variable-length sequence of acoustic features such as Mel-Frequency Cepstral Coefficients~(MFCCs),~$\mathbf{x} = (\mathbf{x}_{1}, \mathbf{x}_{2}, ..., \mathbf{x}_{T})$, where~$\mathbf{x}_{t}$ is the acoustic feature at time~$t$ and~$T$ is the length of the sequence.
We desire that this word embedding is able to describe the semantics of the original audio segment to some degree.
Here we first review the RNN Encoder-Decoder framework, followed by formally proposing the Speech2Vec model.

\begin{figure*}[htp]
  \centering
  \subfigure[Speech2Vec with skipgrams]{\includegraphics[scale=0.32]{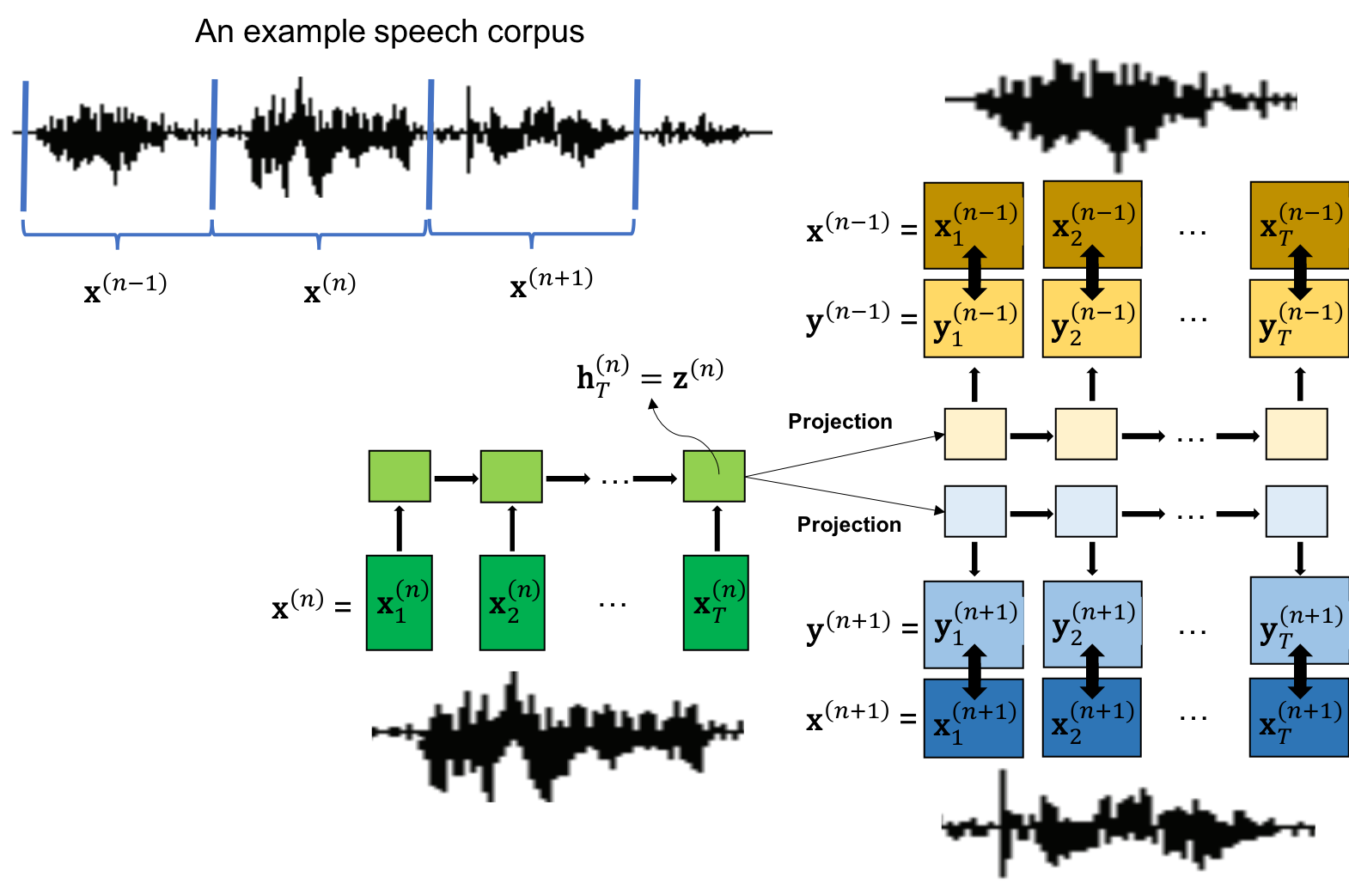} \label{fig:speech2vec-skipgram}}\qquad
  \subfigure[Speech2Vec with cbow]{\includegraphics[scale=0.32]{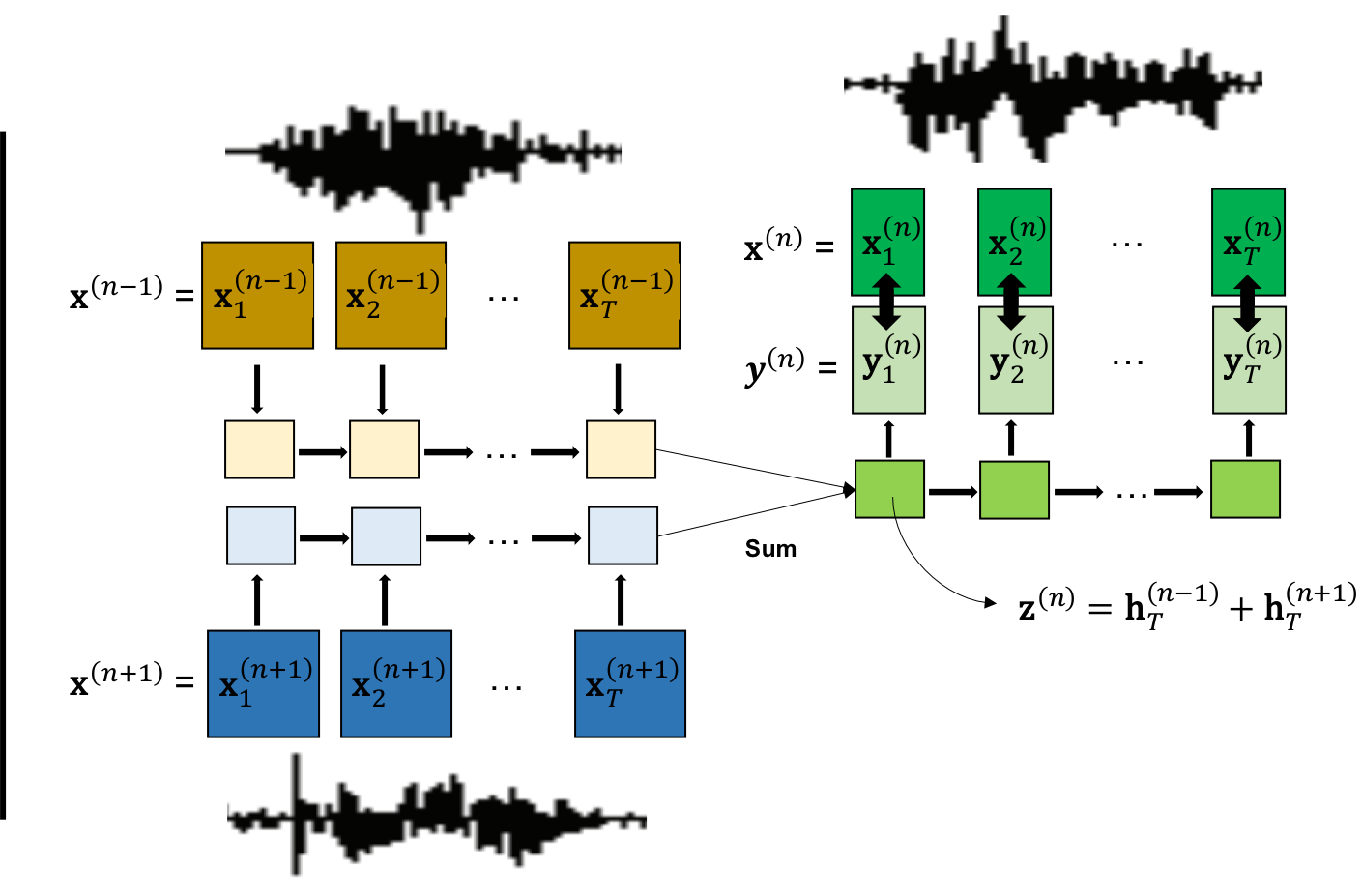} \label{fig:speech2vec-cbow}}
  \caption{
    The structures of Speech2Vec trained with skipgrams and cbow, respectively.
    All audio segments were padded by zero vectors into the same length~$T$.
    Note that when training Speech2Vec with skipgrams, it is the same Decoder RNN that generates all the output audio segments; when training Speech2Vec with cbow, it is the same Encoder RNN that encodes all the input audio segments.
  }
\end{figure*}

\subsection{RNN Encoder-Decoder Framework}
A Recurrent Neural Network~(RNN) Encoder-Decoder consists of an Encoder RNN and a Decoder RNN~\cite{sutskever2014sequence,cho2014learning}.
For an input sequence~$\mathbf{x} = (\mathbf{x}_{1}, \mathbf{x}_{2}, ..., \mathbf{x}_{T})$, the Encoder reads each of its symbol~$\mathbf{x}_{i}$ sequentially, and the hidden state~$\mathbf{h}_{t}$ of the RNN is updated accordingly.
After the last symbol~$\mathbf{x}_{T}$ is processed, the corresponding hidden state~$\mathbf{h}_{T}$ is interpreted as the learned representation of the entire input sequence.
Subsequently, by initializing its hidden state using~$\mathbf{h}_{T}$, the Decoder generates an output sequence~$\mathbf{y} = (\mathbf{y}_{1}, \mathbf{y}_{2}, ..., \mathbf{y}_{T'})$ sequentially, where~$T$ and~$T'$ can be different.
Such a sequence-to-sequence framework does not constrain the input or target sequences, and has been successfully applied to tasks such as machine translation~\cite{luong2015effective,bahdanau2014neural}, video caption generation~\cite{venugopalan2015sequence}, abstract meaning representation parsing and generation~\cite{konstas2017neural}, and acoustic word embeddings acquisition~\cite{chung2016audio}.

\subsection{Speech2Vec}
The backbone of Speech2Vec is the RNN Encoder-Decoder framework.
Inspired by Word2Vec, here we propose two methodologies for training Speech2Vec: skipgrams and continuous bag-of-words~(cbow).
The two variants are depicted in Figure~\ref{fig:speech2vec-skipgram} and Figure~\ref{fig:speech2vec-cbow}, respectively.

\subsubsection{Training Speech2Vec with skipgrams}
The idea of training Speech2Vec with skipgrams is that for each audio segment~$\mathbf{x}^{(n)}$~(corresponding to a word) in a speech corpus, the model is trained to predict the audio segments~$\{\mathbf{x}^{(n - k)}, ..., \mathbf{x}^{(n - 1)}, \mathbf{x}^{(n + 1)}, ..., \mathbf{x}^{(n + k)}\}$~(corresponding to nearby words) within a certain range~$k$ before and after~$\mathbf{x}^{(n)}$.
During training, the Encoder first takes~$\mathbf{x}^{(n)}$
as input and encodes it into a vector representation of fixed dimensionality~$\mathbf{z}^{(n)}$.
The Decoder then maps~$\mathbf{z}^{(n)}$ to several output sequences~$\mathbf{y}^{(i)}, i\in \{n - k, ..., n - 1, n + 1, ..., n + k\}$.
The model is trained by minimizing the gap between the output sequences and their corresponding nearby audio segments, measured by the general mean squared error~$
\sum_{i}\begin{Vmatrix}\mathbf{x}^{(i)} - \mathbf{y}^{(i)}\end{Vmatrix}^2$.
The intuition behind the this approach is that, in order to successfully decode nearby audio segments, the encoded vector representation~$\mathbf{z}^{(n)}$ should contain sufficient semantic information about the current audio segment~$\mathbf{x}^{(n)}$.
After training,~$\mathbf{z}^{(n)}$ is taken as the word embedding of~$\mathbf{x}^{(n)}$.

\subsubsection{Training Speech2Vec with cbow}
In contrast to training Speech2Vec with skipgrams that aim to predict nearby audio segments from~$\mathbf{z}^{(n)}$, training Speech2Vec with cbow sets~$\mathbf{x}^{(n)}$ as the target and aims to infer it from nearby audio segments.
During training, all nearby audio segments are encoded by a shared Encoder into~$\mathbf{h}^{(i)}, i\in \{n - k, ..., n - 1, n + 1, ..., n + k\}$, and their sum~$\mathbf{z}^{(n)} = \sum_{i}\mathbf{h}^{(i)}$ is then used by the Decoder to generate~$\mathbf{x}^{(n)}$.
After training,~$\mathbf{z}^{(n)}$ is taken as the word embedding for~$\mathbf{x}^{(n)}$.
In our experiments, we found that Speech2Vec trained with skipgrams consistently outperforms that trained with cbow.

\subsection{Differences between Speech2Vec and Word2Vec}
\label{sec:speech2vec-vs-word2vec}
The proposed Speech2Vec aims to learn a fixed-length embedding of an audio segment that captures the semantic information of the spoken word directly from audio.
It can be viewed as a speech version of Word2Vec.
Although they have many properties in common, such as sharing the same training methodologies~(skipgrams and cbow), and learning word embeddings that capture semantic information from their respective modalities, it is important to identify two fundamental differences.
First, the architecture of a Word2Vec model is a two-layered fully-connected neural network with one-hot encoded vectors as input and output.
In contrast, the Speech2Vec model is composed of Encoder and Decoder RNNs, in order to handle variable-length input and output sequences of acoustic features.
Second, in a Word2Vec model, the embedding for a particular word is deterministic.
Every instance of the same word will be represented by one, and only one, embedding vector.
In contrast, in the Speech2Vec model, due to the fact that every instance of a spoken word will be different~(due to speaker, channel, and other contextual differences etc.), every instance of the same underlying word will be represented by a {\it different}~(though hopefully similar) embedding vector.
For experimental purposes, in this work, all vectors representing instances of the same spoken word are averaged to obtain a single word embedding.
The effect of this averaging operation is discussed in Section~\ref{sec:exp}.

\begin{table*}[htbp]
  \centering
  \caption{The relationship between the embedding size and the performance. For each of the four models, the highest~$\rho$ of each benchmark is marked in bold. The highest~$\rho$ of each benchmark among all four models is further colored in red.}
  \label{tab:dim-rho}
  \resizebox{\textwidth}{!}{
  \begin{tabular}{ccccc|cccc|cccc|cccc}
    \toprule
    \multirow{2}{*}{Model}  &  \multicolumn{8}{c}{Speech2Vec}  &  \multicolumn{8}{c}{Word2Vec}  \\
    \cmidrule(lr){2-9}  \cmidrule(lr){10-17}
    &  \multicolumn{4}{c}{cbow}  &  \multicolumn{4}{c}{skipgrams}  &  \multicolumn{4}{c}{cbow}  &  \multicolumn{4}{c}{skipgrams}  \\
    \midrule
    Vector dim.  &  10  &  50  &  100  &  200  &  10  &  50  &  100  &  200  &  10  &  50  &  100  &  200  &  10  &  50  &  100  &  200  \\
    \midrule
    \midrule
          Verb-143               &  0.182  &  {\bf 0.223}  &  0.203  &  0.205  &  0.263  &  {\bf 0.315}  &  0.276  &  0.222  &  0.296  &  0.380  &  0.383  &  {\color{red}{\bf 0.385}}  &  0.307  &  0.378  &  {\bf 0.384}  &  0.365  \\
      SimLex-999             &  0.183  &  0.235  &  {\bf 0.238}  &  0.237  &  0.200  &  0.292  &  0.317  &  {\color{red}{\bf 0.335}}  &  0.118  &  {\bf 0.146}  &  0.142  &  0.140  &  0.202  &  0.280  &  0.298  &  {\bf 0.300}  \\
      MC-30                  &  0.680  &  {\bf 0.716}  &  0.688  &  0.684  &  0.701  &  {\color{red}{\bf 0.846}}  &  0.815  &  0.787  &  0.524  &  {\bf 0.539}  &  0.532  &  0.521  &  0.726  &  {\bf 0.762}  &  0.746  &  0.713  \\
      WS-353                 &  0.305  &  {\bf 0.343}  &  0.336  &  0.335  &  0.370  &  {\color{red}{\bf 0.508}}  &  0.502  &  0.498  &  0.198  &  {\bf 0.234}  &  0.228  &  0.233  &  0.334  &  0.452  &  0.455  &  {\bf 0.471}  \\
      WS-353-SIM             &  0.461  &  {\bf 0.484}  &  0.474  &  0.471  &  0.533  &  {\color{red}{\bf 0.663}}  &  0.653  &  0.636  &  0.313  &  {\bf 0.335}  &  0.330  &  0.334  &  0.491  &  0.602  &  0.599  &  {\bf 0.605}  \\
      WS-353-REL             &  0.122  &  {\bf 0.192}  &  0.189  &  0.186  &  0.207  &  {\color{red}{\bf 0.346}}  &  0.332  &  0.331  &  0.051  &  {\bf 0.106}  &  0.095  &  0.100  &  0.172  &  0.308  &  0.308  &  {\bf 0.327}  \\
      RG-65                  &  0.676  &  {\bf 0.705}  &  0.699  &  0.697  &  0.702  &  {\color{red}{\bf 0.790}}  &  0.756  &  0.740  &  0.421  &  0.425  &  0.424  &  {\bf 0.428}  &  0.666  &  {\bf 0.752}  &  0.749  &  0.724  \\
      MEN                    &  0.476  &  {\bf 0.509}  &  0.501  &  0.498  &  0.543  &  {\bf 0.619}  &  0.606  &  0.573  &  0.427  &  {\bf 0.465}  &  0.461  &  0.459  &  0.563  &  0.642  &  {\color{red}{\bf 0.646}}  &  0.632  \\
      MTurk-287              &  0.346  &  {\bf 0.349}  &  0.336  &  0.331  &  0.426  &  {\bf 0.468}  &  0.442  &  0.398  &  0.368  &  0.387  &  {\bf 0.390}  &  0.389  &  0.430  &  {\color{red}{\bf 0.504}}  &  0.503  &  0.469  \\
      MTurk-771              &  0.356  &  {\bf 0.391}  &  0.380  &  0.377  &  0.445  &  {\color{red}{\bf 0.521}}  &  0.503  &  0.463  &  0.246  &  {\bf 0.290}  &  0.289  &  0.288  &  0.413  &  0.499  &  {\bf 0.504}  &  0.479  \\
      SimVerb-3500           &  0.098  &  0.122  &  {\bf 0.126}  &  0.125  &  0.100  &  0.157  &  0.183  &  {\color{red}{\bf 0.204}}  &  0.049  &  {\bf 0.075}  &  0.072  &  0.069  &  0.090  &  0.149  &  0.176  &  {\bf 0.193}  \\
      Rare-Word              &  0.240  &  0.273  &  {\bf 0.275}  &  0.269  &  0.249  &  {\bf 0.323}  &  0.321  &  0.317  &  0.230  &  0.307  &  0.309  &  {\bf 0.310}  &  0.286  &  0.408  &  0.419  &  {\color{red}{\bf 0.431}}  \\
      YP-130                 &  0.198  &  {\bf 0.216}  &  0.211  &  0.214  &  0.322  &  0.321  &  {\bf 0.334}  &  0.302  &  0.231  &  {\bf 0.261}  &  0.257  &  0.253  &  0.345  &  0.391  &  0.431  &  {\color{red}{\bf 0.448}}  \\
    \bottomrule
  \end{tabular}
  }
\end{table*}

\section{Experiments}
\label{sec:exp}
\subsection{Dataset}
For our experiments we used LibriSpeech~\cite{panayotov2015librispeech}, a corpus of read English speech, to learn Speech2Vec embeddings.
In particular, we used a 500 hour subset of broadband speech produced by 1,252 speakers.
Speech features consisting of 13 dimensional Mel Frequency Cepstral Coefficients~(MFCCs) were produced every 10ms.
The speech was segmented according to word boundaries obtained by forced alignment with respect to the reference transcriptions such that each audio segment corresponds to a spoken word.
This resulted in a large set of audio segments~$\{\mathbf{x}^{(1)}, \mathbf{x}^{(2)}, ..., \mathbf{x}^{(|C|)}\}$, where~$|C|$ denotes the total number of audio segments~(words) in the corpus.

\subsection{Implementation and Training Details}
We implemented the Speech2Vec model with PyTorch~\cite{paszke2017automatic}.
The Encoder RNN is a single-layered bidirectional LSTM~\cite{hochreiter1997long}, and the Decoder RNN is another single-layered unidirectional LSTM.
To facilitate the learning process, we also adopted the attention mechanism that enables the Decoder to condition every decoding step on the last hidden state of the Encoder~\cite{subramanian2018learning}, in other words, the Decoder can refer to~$\mathbf{h}_{T}$ when generating every symbol~$\mathbf{y}_{t}$ of the output sequence~$\mathbf{y}$.
The window size~$k$ for training the model with skipgrams and cbow is set to three.
The model was trained by stochastic gradient descent~(SGD) without momentum, with a fixed learning rate of~$1e-3$ and~500 epochs.
We experimented with hyperparameter combinations for training the Speech2Vec model, including the depths of the Encoder and Decoder RNNs, which memory cell~(LSTM or GRU) to use, and  bidirectional or unidirectional RNNs.
We conducted experiments using the specified architecture since it produced the most stable and satisfactory results.

\subsection{Evaluation}
Existing schemes for evaluating methods for word embeddings fall into two major categories: extrinsic and intrinsic~\cite{schnabel2015eval}.
With the extrinsic method, the learned word embeddings are used as input features to a downstream task~\cite{yu2017character,lample2016neural,plank2016multilingual,kim2016character,ballesteros2015improved}, and the performance metric varies from task to task.
The intrinsic method directly tests for semantic or syntactic relationships between words, and includes the tasks of word similarity and word analogy~\cite{mikolov2013distributed}.
In this paper, we focus on the intrinsic method, especially the word similarity task, for evaluating and analyzing the Speech2Vec word embeddings.

We used 13 benchmarks~\cite{faruqui2014community} to measure word similarity, including \textbf{WS-353}~\cite{yang2006verb}, \textbf{WS-353-REL}~\cite{agirre2009study}, \textbf{WS-353-SIM}, \textbf{MC-30}~\cite{miller1991contextual}, \textbf{RG-65}~\cite{rubenstein1965contextual}, \textbf{Rare-Word}~\cite{luong2013better}, \textbf{MEN}~\cite{bruni2012distributional}, \textbf{MTurk-287}~\cite{radinsky2011word}, \textbf{MTurk-771}~\cite{halawi2012large}, \textbf{YP-130}~\cite{yang2006verb}, \textbf{SimLex-999}~\cite{hill2015simlex}, \textbf{Verb-143}~\cite{baker2014unsupervised}, and \textbf{SimVerb-3500}~\cite{gerz2016simverb}.
These~13 benchmarks contain different numbers of pairs of English words that have been assigned similarity ratings by humans, and each of them evaluates the word embeddings in terms of different aspects.
For example, \textbf{RG-65} and \textbf{MC-30} focus on nouns, \textbf{YC-130} and \textbf{SimVerb-3500} focus on verbs, and \textbf{Rare-Word} focuses on rare-words.
The similarity between a given pair of words was calculated by computing the cosine similarity between their corresponding word embeddings.
We then reported the Spearman's rank correlation coefficient~$\rho$ between the rankings produced by each model against the human rankings~\cite{myers1995research}.

We compared Speech2Vec trained with skipgrams or cbow with its Word2Vec counterpart trained on the transcriptions of the LibriSpeech corpus using the fastText implementation~\cite{bojanowski2017enriching}.
For convenience, we refer to these four models as skipgrams Speech2Vec, cbow Speech2Vec, skipgrams Word2Vec, and cbow Word2Vec, respectively.

\begin{table*}[htbp]
  \centering
  \caption{The relationship between the size of the training corpus and the performance. The percentage denotes the proportion of the entire corpus that was used for training the models. The reported results are based on the word embeddings of 50-dim.}
  \label{tab:trainSize-rho}
  \resizebox{\textwidth}{!}{
  \begin{tabular}{ccccc|cccc|cccc|cccc}
    \toprule
    \multirow{2}{*}{Model}  &  \multicolumn{8}{c}{Speech2Vec}  &  \multicolumn{8}{c}{Word2Vec}  \\
    \cmidrule(lr){2-9}  \cmidrule(lr){10-17}
    &  \multicolumn{4}{c}{cbow}  &  \multicolumn{4}{c}{skipgrams}  &  \multicolumn{4}{c}{cbow}  &  \multicolumn{4}{c}{skipgrams}  \\
    \midrule
      Training size          &   10\%   &   40\%   &   70\%  &   100\%  &   10\%  &   40\%  &   70\%  &  100\%  &   10\%   &   40\%   &   70\%  &  100\%  &   10\%  &   40\%  &   70\%  &   100\%  \\
      \midrule
      \midrule
      Verb-143               &  0.090   &  0.071   &  0.116  &  0.223  &  0.098   &  0.152  &  0.220  &  0.315  &  0.196   &  0.257   &  0.331  &  0.380  &  0.148  &  0.259  &  0.328  &  0.378  \\
      WS-353                 &  -0.101  &  0.211   &  0.319  &  0.343  &  0.066   &  0.392  &  0.459  &  0.508  &  0.045   &  0.091   &  0.167  &  0.234  &  0.129  &  0.377  &  0.412  &  0.452  \\
      RG-65                  &  0.024   &  0.199   &  0.593  &  0.705  &  0.020   &  0.605  &  0.661  &  0.790  &  0.196   &  0.192   &  0.333  &  0.425  &  0.330  &  0.416  &  0.642  &  0.752  \\
      MEN                    &  0.033   &  0.311   &  0.451  &  0.509  &  0.283   &  0.506  &  0.585  &  0.619  &  0.016   &  0.258   &  0.403  &  0.465  &  0.247  &  0.541  &  0.621  &  0.642  \\
      MTurk-771              &  0.098   &  0.246   &  0.321  &  0.391  &  0.186   &  0.416  &  0.462  &  0.521  &  0.094   &  0.148   &  0.223  &  0.290  &  0.182  &  0.392  &  0.474  &  0.499  \\
      YP-130                 &  -0.027  &  0.067   &  0.181  &  0.216  &  0.097   &  0.196  &  0.311  &  0.321  &  0.064   &  0.085   &  0.182  &  0.256  &  0.403  &  0.216  &  0.365  &  0.391  \\
    \bottomrule
  \end{tabular}
  }
\end{table*}

\subsection{Results and Discussions}
We trained the four models with different embedding sizes to understand how large the embedding size should be to capture sufficient semantic information about the word.
The results are shown in Table~\ref{tab:dim-rho}.
We also varied the size of the corpus used for training the four models and report the results in Table~\ref{tab:trainSize-rho}.%
\footnote{For Table~\ref{tab:trainSize-rho}, we randomly picked six benchmarks to report due to page limits, but the conclusion remains the same.}
The numbers in both tables are the average of running the experiment 10 times and the standard deviations are negligible.
From Table~\ref{tab:dim-rho} and Table~\ref{tab:trainSize-rho}, we have the following observations.

\newcommand{\myparagraph}[1]{\vspace{.4em} \noindent \textbf{#1}\ }
\myparagraph{Embedding size impact on performance.}
We found that increasing the embedding size does not always result in improved performance.
For cbow Speech2Vec, skipgrams Speech2Vec, and cbow Word2Vec, word embeddings of 50-dimensions are able to capture enough semantic information of the words, as the best performance~(highest~$\rho$) of each benchmark is mostly achieved by them.
For skipgrams Word2Vec, although the best performance of 7 out of 13 benchmarks is achieved by word embeddings of 200-dims, there are 6 benchmarks whose best performance is achieved by word embeddings of other sizes.

\myparagraph{Comparing Speech2Vec to Word2Vec.}
From Table~\ref{tab:dim-rho} we see that skipgrams Speech2Vec achieves the highest~$\rho$ in 8 out of 13 benchmarks, outperforming cbow and skipgrams Word2Vec in combination.
We believe a possible reason for such results is due to skipgrams Speech2Vec's ability to capture semantic information present in speech such as prosody that is not in text.

\myparagraph{Comparing skipgrams to cbow Speech2Vec.}
From Table~\ref{tab:dim-rho} we observe that skipgrams Speech2Vec consistently outperforms cbow Speech2Vec on all benchmarks for all embedding sizes.
This result aligns with the empirical fact that skipgrams Word2Vec is likely to work better than cbow Word2Vec with small training corpus size~\cite{mikolov2013distributed}.

\myparagraph{Impact of training corpus size.}
From Table~\ref{tab:trainSize-rho} we observe that when 10\% of the corpus was used for training, the resulting word embeddings perform poorly.
Unsurprisingly, the performance continues to improve as training size increases.

\subsection{Variance Study}
In Section~\ref{sec:speech2vec-vs-word2vec} we mention that in Speech2Vec, every instance of a spoken word will produce a different embedding vector.
Here we try to understand how the vectors for a given word vary. i.e., are they similar, or is there considerable variance that the averaging operation adopted in this paper smooths out?

\begin{figure}[htbp]
  \centering
  \includegraphics[scale=0.38]{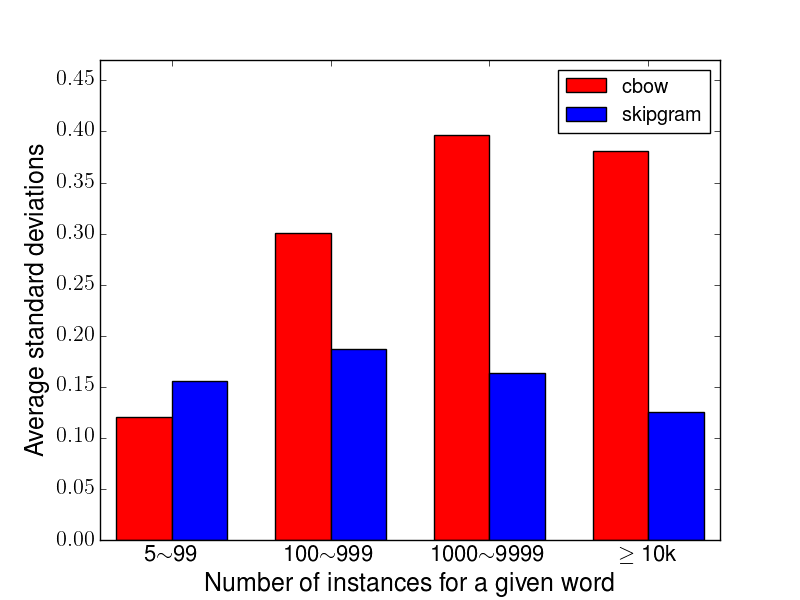}
  \caption{How the vector representations for a given word vary with respect to the times it appears in the corpus.}
  \label{fig:std-mean-analysis}
\end{figure}

To study this, we partitioned all words into four sub-groups based on the number of times,~$N$, that they appeared in the corpus, ranging from ~$5\sim 99, 100\sim 999, 1000\sim 9999,\text{and} \geq 10k$.
Then, for all vector representations~$\{\mathbf{w}^{1}, \mathbf{w}^{2}, ..., \mathbf{w}^{N}\}$ of a given word~$w$ that appeared~$N$ times, we computed the mean of the standard deviations of each dimensions~$m_{w} = \frac{1}{d}\sum_{i = 1}^{d}{\tt std}(\mathbf{w}^{1}, \mathbf{w}^{2}, ..., \mathbf{w}^{N})$, where~$d$ denotes the embedding size.
Finally, we averaged~$m_{w}$ for every word~$w$ that belongs to the same sub-group and reported the results in Figure~\ref{fig:std-mean-analysis}.
\vspace{-0.7cm}

From Figure~\ref{fig:std-mean-analysis} we observe that when~$N$ falls in~$5\sim 99$, the variances of the vectors generated by cbow Speech2Vec are smaller than those generated by skipgrams Speech2Vec.
However, when~$N$ becomes bigger, variances of the vectors generated by skipgrams Speech2Vec become smaller than those generated by cbow Speech2Vec, and the gap continues to grow as~$N$ increases.
We suspect the lower variation of the skipgrams model relative to the cbow model is related to the overall superior performance of the skipgrams Speech2Vec model.
We are encouraged that the deviation of the skipgrams model gets smaller as~$N$ increases, as it suggests stability in the model.

\subsection{Visualizations}
\begin{figure}[htp]
  \centering
  \includegraphics[scale=0.38]{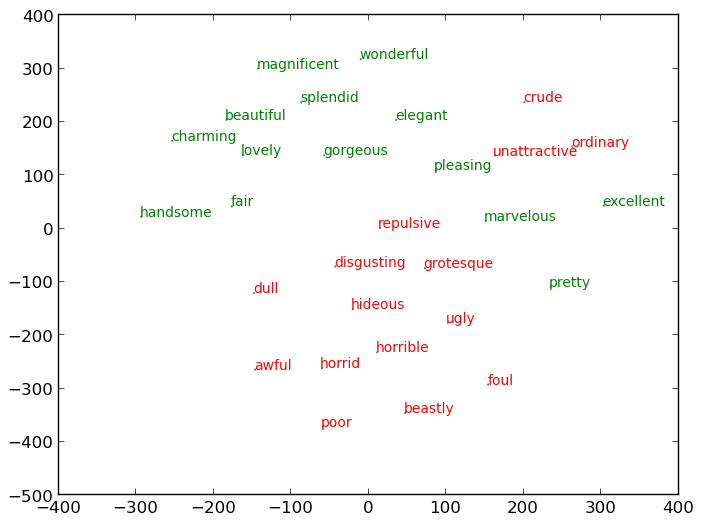}
  \caption{t-SNE projection of the word embeddings learned by skipgrams Speech2Vec. Words with positive and negative meanings were colored in green and red, respectively.}
  \label{fig:visualize-skipgrams-speech2vec}
\end{figure}
\vspace{-0.2cm}
We visualized the word embeddings learned by skipgrams Speech2Vec with t-SNE~\cite{maaten2008visualizing} using~\url{http://www.wordvectors.org/} in Figure~\ref{fig:visualize-skipgrams-speech2vec}.
We see that words with positive meanings~(colored in green) are mainly located at the upper part of the figure, while words with negative meanings~(colored in red) are mostly located at the bottom.
Such distribution suggests that the learned word embeddings do capture notions of antonym and synonyms to some degree.



\section{Conclusions and Future Work}
Speech2Vec, which integrates a RNN Encoder-Decoder framework with skipgrams or cbow for training, extends the text-based Word2Vec~\cite{mikolov2013distributed} model to learn word embeddings directly from speech.
Speech2Vec has access to richer information in the speech signal that does not exist in plain text.
In our experiments, the learned word embeddings outperform those produced by Word2Vec from the transcriptions.
In the future, we plan to evaluate the word embeddings on speech-related extrinsic tasks such as machine listening comprehension~\cite{chung2018supervised,tseng2016towards} and speech-based visual question answering~\cite{zhang2017speech} by initializing the embedding layers of the neural network models.
Finally, in this work, some supervision was incorporated into the learning by using forced alignment segmentations as the basis for audio segments.
It would be interesting to explore less supervised segmentations to learn word boundaries~\cite{kamper2017learning,kamper2017segmental}.

\bibliographystyle{IEEEtran}
\bibliography{mybib}

\begin{thebibliography}{10}
\providecommand{\url}[1]{#1}
\csname url@samestyle\endcsname
\providecommand{\newblock}{\relax}
\providecommand{\bibinfo}[2]{#2}
\providecommand{\BIBentrySTDinterwordspacing}{\spaceskip=0pt\relax}
\providecommand{\BIBentryALTinterwordstretchfactor}{4}
\providecommand{\BIBentryALTinterwordspacing}{\spaceskip=\fontdimen2\font plus
\BIBentryALTinterwordstretchfactor\fontdimen3\font minus
  \fontdimen4\font\relax}
\providecommand{\BIBforeignlanguage}[2]{{%
\expandafter\ifx\csname l@#1\endcsname\relax
\typeout{** WARNING: IEEEtran.bst: No hyphenation pattern has been}%
\typeout{** loaded for the language `#1'. Using the pattern for}%
\typeout{** the default language instead.}%
\else
\language=\csname l@#1\endcsname
\fi
#2}}
\providecommand{\BIBdecl}{\relax}
\BIBdecl

\bibitem{mikolov2013distributed}
T.~Mikolov, I.~Sutskever, K.~Chen, G.~S. Corrado, and J.~Dean, ``Distributed
  representations of words and phrases and their compositionality,'' in
  \emph{NIPS}, 2013.

\bibitem{bojanowski2017enriching}
P.~Bojanowski, E.~Grave, A.~Joulin, and T.~Mikolov, ``Enriching word vectors
  with subword information,'' \emph{Transactions of the Association for
  Computational Linguistics}, vol.~5, pp. 135--146, 2017.

\bibitem{pennington2014glove}
J.~Pennington, R.~Socher, and C.~Manning, ``Glove: Global vectors for word
  representation,'' in \emph{EMNLP}, 2014.

\bibitem{yu2017character}
X.~Yu and N.~T. Vu, ``Character composition model with convolutional neural
  networks for dependency parsing on morphologically rich languages,'' in
  \emph{ACL}, 2017.

\bibitem{lample2016neural}
G.~Lample, M.~Ballesteros, S.~Subramanian, K.~Kawakami, and C.~Dyer, ``Neural
  architectures for named entity recognition,'' in \emph{NAACL HLT}, 2016.

\bibitem{plank2016multilingual}
B.~Plank, A.~S{\o}gaard, and Y.~Goldberg, ``Multilingual part-of-speech tagging
  with bidirectional long short-term memory models and auxiliary loss,'' in
  \emph{ACL}, 2016.

\bibitem{kim2016character}
Y.~Kim, Y.~Jernite, D.~Sontag, and A.~M. Rush, ``Character-aware neural
  language models,'' in \emph{AAAI}, 2016.

\bibitem{ballesteros2015improved}
M.~Ballesteros, C.~Dyer, and N.~A. Smith, ``Improved transition-based parsing
  by modeling characters instead of words with {LSTMs},'' in \emph{EMNLP},
  2015.

\bibitem{he2017multi}
W.~He, W.~Wang, and K.~Livescu, ``Multi-view recurrent neural acoustic word
  embeddings,'' in \emph{ICLR}, 2017.

\bibitem{settle2016discriminative}
S.~Settle and K.~Livescu, ``Discriminative acoustic word embeddings: Recurrent
  neural network-based approaches,'' in \emph{SLT}, 2016.

\bibitem{chung2016audio}
Y.-A. Chung, C.-C. Wu, C.-H. Shen, H.-Y. Lee, and L.-S. Lee, ``Audio word2vec:
  Unsupervised learning of audio segment representations using
  sequence-to-sequence autoencoder,'' in \emph{INTERSPEECH}, 2016.

\bibitem{kamper2016deep}
H.~Kamper, W.~Wang, and K.~Livescu, ``Deep convolutional acoustic word
  embeddings using word-pair side information,'' in \emph{ICASSP}, 2016.

\bibitem{bengio2014word}
S.~Bengio and G.~Heigold, ``Word embeddings for speech recognition,'' in
  \emph{INTERSPEECH}, 2014.

\bibitem{levin2013fixed}
K.~Levin, K.~Henry, A.~Jansen, and K.~Livescu, ``Fixed-dimensional acoustic
  embeddings of variable-length segments in low-resource settings,'' in
  \emph{ASRU}, 2013.

\bibitem{harwath2017learning}
D.~Harwath and J.~Glass, ``Learning word-like units from joint audio-visual
  analysis,'' in \emph{ACL}, 2017.

\bibitem{harwath2016unsupervised}
D.~Harwath, A.~Torralba, and J.~Glass, ``Unsupervised learning of spoken
  language with visual context,'' in \emph{NIPS}, 2016.

\bibitem{harwath2015deep}
D.~Harwath and J.~Glass, ``Deep multimodal semantic embeddings for speech and
  images,'' in \emph{ASRU}, 2015.

\bibitem{sutskever2014sequence}
I.~Sutskever, O.~Vinyals, and Q.~V. Le, ``Sequence to sequence learning with
  neural networks,'' in \emph{NIPS}, 2014.

\bibitem{cho2014learning}
K.~Cho, B.~van Merri{\"{e}}nboer, {\c C}.~G{\"{u}}l{\c c}ehre, D.~Bahdanau,
  F.~Bougares, H.~Schwenk, and Y.~Bengio, ``Learning phrase representations
  using rnn encoder--decoder for statistical machine translation,'' in
  \emph{EMNLP}, 2014.

\bibitem{chung2017learning}
Y.-A. Chung and J.~Glass, ``Learning word embeddings from speech,'' in
  \emph{NIPS ML4Audio Workshop}, 2017.

\bibitem{luong2015effective}
T.~Luong, H.~Pham, and C.~D. Manning, ``Effective approaches to attention-based
  neural machine translation,'' in \emph{EMNLP}, 2015.

\bibitem{bahdanau2014neural}
D.~Bahdanau, K.~Cho, and Y.~Bengio, ``Neural machine translation by jointly
  learning to align and translate,'' in \emph{ICLR}, 2014.

\bibitem{venugopalan2015sequence}
S.~Venugopalan, M.~Rohrbach, J.~Donahue, R.~Mooney, T.~Darrell, and K.~Saenko,
  ``Sequence to sequence-video to text,'' in \emph{CVPR}, 2015.

\bibitem{konstas2017neural}
I.~Konstas, S.~Iyer, M.~Yatskar, Y.~Choi, and L.~Zettlemoyer, ``Neural amr:
  Sequence-to-sequence models for parsing and generation,'' in \emph{ACL},
  2017.

\bibitem{panayotov2015librispeech}
V.~Panayotov, G.~Chen, D.~Povey, and S.~Khudanpur, ``Librispeech: an {ASR}
  corpus based on public domain audio books,'' in \emph{ICASSP}, 2015.

\bibitem{paszke2017automatic}
A.~Paszke, S.~Gross, S.~Chintala, G.~Chanan, E.~Yang, Z.~DeVito, Z.~Lin,
  A.~Desmaison, L.~Antiga, and A.~Lerer, ``Automatic differentiation in
  pytorch,'' in \emph{NIPS Autodiff Workshop}, 2017.

\bibitem{hochreiter1997long}
S.~Hochreiter and J.~Schmidhuber, ``Long short-term memory,'' \emph{Neural
  Computation}, vol.~9, no.~8, pp. 1735--1780, 1997.

\bibitem{subramanian2018learning}
S.~Subramanian, A.~Trischler, Y.~Bengio, and C.~Pal, ``Learning general purpose
  distributed sentence representations via large scale multi-task learning,''
  in \emph{ICLR}, 2018.

\bibitem{schnabel2015eval}
T.~Schnabel, I.~Labutov, D.~Mimno, and T.~Joachims, ``Evaluation methods for
  unsupervised word embeddings,'' in \emph{EMNLP}, 2015.

\bibitem{faruqui2014community}
M.~Faruqui and C.~Dyer, ``Community evaluation and exchange of word vectors at
  wordvectors.org,'' in \emph{ACL System Demonstrations}, 2014.

\bibitem{yang2006verb}
D.~Yang and D.~M. Powers, ``Verb similarity on the taxonomy of wordnet,'' in
  \emph{GWC}, 2006.

\bibitem{agirre2009study}
E.~Agirre, E.~Alfonseca, K.~Hall, J.~Kravalova, M.~Pa{\c{s}}ca, and A.~Soroa,
  ``A study on similarity and relatedness using distributional and
  wordnet-based approaches,'' in \emph{NAACL HLT}, 2009.

\bibitem{miller1991contextual}
G.~A. Miller and W.~G. Charles, ``Contextual correlates of semantic
  similarity,'' \emph{Language and Cognitive Processes}, vol.~6, no.~1, pp.
  1--28, 1991.

\bibitem{rubenstein1965contextual}
H.~Rubenstein and J.~B. Goodenough, ``Contextual correlates of synonymy,''
  \emph{Communications of the ACM}, vol.~8, no.~10, pp. 627--633, 1965.

\bibitem{luong2013better}
M.-T. Luong, R.~Socher, and C.~D. Manning, ``Better word representations with
  recursive neural networks for morphology,'' in \emph{CoNLL}, 2013.

\bibitem{bruni2012distributional}
E.~Bruni, G.~Boleda, M.~Baroni, and N.-K. Tran, ``Distributional semantics in
  technicolor,'' in \emph{ACL}, 2012.

\bibitem{radinsky2011word}
K.~Radinsky, E.~Agichtein, E.~Gabrilovich, and S.~Markovitch, ``A word at a
  time: computing word relatedness using temporal semantic analysis,'' in
  \emph{WWW}, 2011.

\bibitem{halawi2012large}
G.~Halawi, G.~Dror, E.~Gabrilovich, and Y.~Koren, ``Large-scale learning of
  word relatedness with constraints,'' in \emph{KDD}, 2012.

\bibitem{hill2015simlex}
F.~Hill, R.~Reichart, and A.~Korhonen, ``Simlex-999: Evaluating semantic models
  with (genuine) similarity estimation,'' \emph{Computational Linguistics},
  vol.~41, no.~4, pp. 665--695, 2015.

\bibitem{baker2014unsupervised}
S.~Baker, R.~Reichart, and A.~Korhonen, ``An unsupervised model for instance
  level subcategorization acquisition,'' in \emph{EMNLP}, 2014.

\bibitem{gerz2016simverb}
D.~Gerz, I.~Vuli{\'{c}}, F.~Hill, R.~Reichart, and A.~Korhonen, ``Simverb-3500:
  A large-scale evaluation set of verb similarity,'' in \emph{EMNLP}, 2016.

\bibitem{myers1995research}
J.~L. Myers and A.~D. Well, \emph{Research design and statistical analysis},
  1st~ed.\hskip 1em plus 0.5em minus 0.4em\relax Routledge, 6 1995.

\bibitem{maaten2008visualizing}
L.~v.~d. Maaten and G.~Hinton, ``Visualizing data using t-sne,'' \emph{Journal
  of Machine Learning Research}, vol.~9, no. Nov, pp. 2579--2605, 2008.

\bibitem{chung2018supervised}
Y.-A. Chung, H.-Y. Lee, and J.~Glass, ``Supervised and unsupervised transfer
  learning for question answering,'' in \emph{NAACL HLT}, 2018.

\bibitem{tseng2016towards}
B.-H. Tseng, S.-S. Shen, H.-Y. Lee, and L.-S. Lee, ``Towards machine
  comprehension of spoken content: Initial {TOEFL} listening comprehension test
  by machine,'' in \emph{INTERSPEECH}, 2016.

\bibitem{zhang2017speech}
T.~Zhang, D.~Dai, T.~Tuytelaars, M.-F. Moens, and L.~Van~Gool, ``Speech-based
  visual question answering,'' \emph{CoRR}, vol. abs/1705.00464, 2017.

\bibitem{kamper2017learning}
H.~Kamper, K.~Livescu, and S.~Goldwater, ``An embedded segmental k-means model
  for unsupervised segmentation and clustering of speech,'' in \emph{ASRU},
  2017.

\bibitem{kamper2017segmental}
H.~Kamper, A.~Jansen, and S.~Goldwater, ``A segmental framework for
  fully-unsupervised large-vocabulary speech recognition,'' \emph{Computer
  Speech and Language}, vol.~46, no.~C, pp. 154--174, 2017.

\end{thebibliography}

\end{document}